\documentclass[10pt,twocolumn,letterpaper]{article}

\usepackage{cvpr}              % To produce the CAMERA-READY version
\usepackage{multirow}
\usepackage{makecell}

\usepackage{float} % 加载float宏包以支持H选项
\usepackage{graphicx} % 加载graphicx宏包以插入图片
\usepackage{xcolor}
\definecolor{cvprblue}{rgb}{0.21,0.49,0.74}
\usepackage[pagebackref,breaklinks,colorlinks,allcolors=cvprblue]{hyperref}

\usepackage{setspace}

\def\paperID{12517} % *** Enter the Paper ID here
\def\confName{CVPR}
\def\confYear{2025}

\def\yhm{\textcolor{black}}

\title{Beyond Background Shift: \\ Rethinking Instance Replay in Continual Semantic Segmentation}

\author{
Hongmei Yin$^{1}$\thanks{Co-first authors. Equal contribution.}~~~~~~~~~Tingliang Feng$^{1}$\footnote[1]~~~~~~~~~Fan Lyu$^2$~~~~~~~~~Fanhua Shang$^1\thanks{Fanhua Shang is the corresponding author of this paper.}$\\
Hongying Liu$^{1}$~~~~~~~~~Wei Feng$^1$~~~~~~~~~Liang Wan$^{1}$\\
{$^1$College of Intelligence and Computing,~Tianjin University}\\
{$^2$New Laboratory of Pattern Recognition, Institute of Automation, Chinese Academy of Sciences}\\
{\tt\small \{hongmeiyin, fengtl, hyliu2009, wfeng, lwan, fhshang\}@tju.edu.cn~~~fan.lyu@cripac.ia.ac.cn}\\
}

\begin{document}
\maketitle
\begin{abstract}
\begin{spacing}{0.96}

In this work, we focus on continual semantic segmentation (CSS), where segmentation networks are required to continuously learn new classes without erasing knowledge of previously learned ones. 
Although storing images of old classes and directly incorporating them into the training of new models has proven effective in mitigating catastrophic forgetting in classification tasks, this strategy presents notable limitations in CSS. 
Specifically, the stored and new images with partial category annotations leads to confusion between unannotated categories and the background, complicating model fitting.
To tackle this issue, this paper proposes a novel Enhanced Instance Replay (EIR) method, which not only preserves knowledge of old classes while simultaneously eliminating background confusion by instance storage of old classes, but also mitigates background shifts in the new images by integrating stored instances with new images. 
By effectively resolving background shifts in both stored and new images, EIR alleviates catastrophic forgetting in the CSS task, thereby enhancing the model's capacity for CSS.
Experimental results validate the efficacy of our approach, which significantly outperforms state-of-the-art CSS methods. The code is available at \url{https://github.com/YikeYin97/EIR}.

\vspace{-0.05in}
\end{spacing}
\end{abstract}

% 尽管存储旧类的图像并将其直接合并到新模型的训练中已被证明在减轻连续学习分类任务中的灾难性遗忘方面是有效的，但将其应用于CSS中时，存储图像和新类图像中因缺失部分类别标注而导致的背景偏移问题，会影响模型对这些类别的识别性能。
% Although storing images of old classes and directly incorporating them into the training of a new model has proven effective in mitigating catastrophic forgetting in continual learning classification tasks, when applied to CSS, the issue of background shift arising from missing partial category annotations in both stored and new class images can adversely affect the model's ability to recognize these categories.

% 为了解决这一障碍，本文提出了ISF，它不仅通过利用存储instance替代存储图像，从而在保留了旧类知识的同时，抑制了存储数据中的背景偏移影响，而且根据新类图像中的上下文信息，将存储instance与新类图像进行融合，从而缓解了新类数据中的背景偏移影响。
% To address this challenge, this paper proposes the \emph{Instance Storage Fusion}, which not only substitutes image storage with instance storage, preserving knowledge of old classes while mitigating background bias in the stored data, but also integrates stored instances with new class images based on contextual information, alleviating the background bias in the new class data.

% 通过有效地解决存储图像和新图像中的背景偏移问题，我们的方法有效地抑制了CSS任务中存在的灾难性遗忘，增强了模型对新旧类别的拟合能力。
% By effectively addressing the background shift between stored and new images, our approach successfully mitigates catastrophic forgetting in the CSS task, thereby enhancing the model's ability to adapt to both new and old classes.    
\section{Introduction}
\label{sec:intro}
\begin{spacing}{0.96}
As the development of continual learning~\cite{castro2018end,chaudhry2019tiny,cauwenberghs2000incremental,wu2019large}, continual semantic segmentation (CSS) was proposed to conduct segmentation in class-incremental scenarios, where the model needs to learn and recognize new classes at the pixel level. However, CSS suffers from the issue of \textit{background shift}, where the background class continuously shifts over the learning step, exacerbating catastrophic forgetting~\cite{cermelli2020modeling,belouadah2019il2m,douillard2020podnet,hou2019learning} that the model under-fits prior knowledge, \ie, older classes.
Generally, background shift roots in the fact that only classes being learned in the current step are correctly labeled, while other old and future classes are mislabeled as background, \ie, \textit{the partial labeling issue}. 

%as shown in Fig.\ref{fig:1} (a).  11/11
% Such mislabeling will confuse the model and make it more difficult to distinguish different unseen classes.

%exacerbating the issue of catastrophic forgetting  making CSS more challenging.

\begin{figure}[t!]
	\centering	\includegraphics[width=1.0\linewidth]{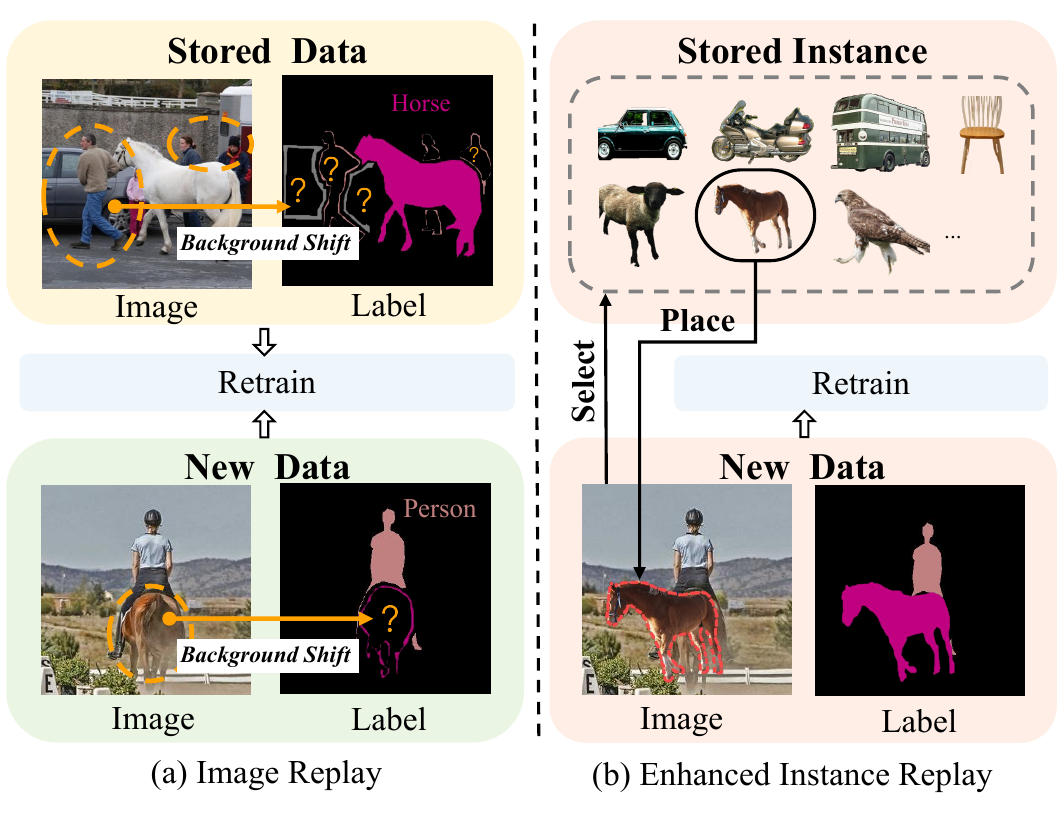}
	\vspace{-0.3in}
	\caption{Comparison of traditional image replay (a) and our replay methods (b). (a) shows that only the old class ``horse'' is labeled in stored image, while other classes (new class ``person'' and old class ``car'') are labeled as background. Both old (``horse'') and future classes in new images are labeled as background. (b) Our method avoids confusing information in stored image by retaining instances and alleviates background shift by fusing these instances in the new image.}
	\label{fig:1}
	\vspace{-0.25in}
\end{figure}

% {Replay} is a well-proved strategy to mitigate catastrophic forgetting in continual learning, which uses a small memory buffer to store and retrain the raw images from the old classes continuously~\cite{robins1995catastrophic,kemker2017fearnet,shin2017continual,castro2018end,chaudhry2019tiny}. 
% \lyu{(Why the reply strategy is good? list reasons or advantages to strengthen motivation.)}
% However, \textit{replay at the image level is ineffective in CSS due to the limitation of partial labeling}. 
% Specifically, the stored images lack annotations for both \lyu{new and other old classes} in CSS. 
% As Fig.\ref{fig:1} (b) shows, only the old class 'horse' is correctly labeled, while the new class 'person' and other old class 'car', are mislabeled as background. 
% Consequently, the model struggles to learn accurate features for these new and other old classes from the stored images, increasing the risk of misclassifying them as background. 

{Replay} is a well-proved strategy to mitigate catastrophic forgetting in general continual learning, \ie, on image classification, which uses a small memory buffer to store and retrain raw images from the old classes continuously~\cite{robins1995catastrophic,kemker2017fearnet,shin2017continual,castro2018end,chaudhry2019tiny,sun2022exploring}. Image replay enables the model to directly access old data during fine-tuning on new data, reinforcing existing knowledge and improving retention of old information.
However, \textit{replay at the image level is ineffective in CSS due to the limitation of partial labeling}.  Specifically, the stored images are also only annotated with partial old classes, while other classes might be present but lack annotation. As Fig.\ref{fig:1} (a) shows, only the old class ``horse'' is correctly labeled, while the new class ``person'' and other old class ``car'', are mislabeled as background. This absence of annotation may disrupt the model's mapping relationship for these various classes, thereby heightening the risk of misclassifying these classes as background.

%11-13 因此，不难发现，部分标注挑战之所以产生，是因为图像重放将多个实例的组合（即图像本身）视为一个整体，而这种固定的组合限制了重放的标注空间，即只有部分旧类有标注。\textit{实际上，一个实例只有一个标签，并且像素数量远少于一张图像，那么我们能否从图像中提取实例并进行重放呢？}我们假设，由于实例不包含令人困惑的背景信息，因此重放实例可以摆脱部分标注的挑战。 遗憾的是，我们发现在Instance replay中，naive的instance replay does not yield satisfactory results。详细来说，简单的将实例作为训练数据输入模型，就像它们是图像回放的简单复制一样，无法设置合适的类别动态组合，呼应类间关系。而随机的复制粘贴方式由于其固有的随机性，无法使得instance有效融入背景中，容易突兀或者污染原本的训练样本,导致训练不稳定。  因此，我们发现，在CSS任务中，instance replay最大的问题就是 类别动态实例组合问题以及背景融合问题。
Therefore, it is easy to find that the partial label challenge happens because the image replay treats the combination of multiple instances as a whole, \ie the image itself, and the fixed combinations limit the label space of the replay. \textit{In fact, an instance has only one label and much fewer pixels than an image, so can we extract instances from images and replay them instead?} We assume that replaying instances gets rid of the partial label challenge since instances do not contain confusing background information. 
Regrettably, naive approaches to instance replay can not yield satisfactory results. Specifically, simply inputting instances into the model for training, as if they were a mere replication of image replay, fails to establish appropriate dynamic combinations of classes that reflect inter-class relationships. 
Additionally, randomly copy-pasting instances into training images prevents instances from effectively integrating into the background, which may result in abruptness or contamination of the original training samples, and leading to unstable training.
Therefore, we argue that in CSS tasks, the two significant issues with instance replay are the dynamic combination of classes and background integration.

% Regrettably, simply inputting instances into the model for training, as if they were a mere replication of image replay, does not yield satisfactory results. \lyu{The reason is that this direct approach overlooks the importance of relationships among classes and between each class and the background in semantic segmentation, which are crucial for reducing confusion. Additionally, it does not mitigate the background shift issue in new data.}

% 一种：random   共有的问题  explore 在CSS问题中  合适的replay方法？
% 最大的问题，我们提出了什么？ 存取贴用，分别是怎么做的？

%11-13 受此启发，本文旨在为持续语义分割（CSS）任务探寻一种更为适合的实例重放策略，以全面考量动态实例组合与背景融合之难题。我们特此提出增强型实例重放（Enhanced Instance Replay, EIR）策略，该策略专为CSS任务设计，成效显著。EIR通过旧类实例覆盖新类图像中的未标注区域，有效减轻新类数据中的背景偏移现象。进一步地，EIR采纳潜在类别预测策略，实现自适应的动态类别组合，确保实例融入图像前后，新图像中的类别上下文保持一致性。同时，EIR在将实例以增强方式融入新图像时，还优化实例在图像中的重放位置，规避了因随机放置而导致的语义不合理的重叠。此外，本研究还引入了一种区域特定的知识蒸馏损失（Region-Specific Knowledge Distillation, RSKD），以更高效地利用已融合至新图像中的旧类实例信息，确保新旧模型在新图像中的旧实例区域上能够给出一致的预测结果，同时避免了对新类别可塑性的不利干扰。 

Motivated by this, this work is to explore a feasible instance replay practice for CSS task, fully accounting for dynamic instance combinations and background integration.
Specifically, we propose a novel Enhanced Instance Replay (EIR) method, which is an effective replay strategy for the CSS task.
First, EIR utilizes old-class instances to cover unannotated areas in new images, thereby mitigating the issue of background shift in new data.
Second, EIR is combined with a latent class prediction strategy to achieve adaptive class combinations, ensuring consistency in the class context of the new image before and after instance integration. 
Third, EIR finds better positions of instances within the new image when integrating them in an enhanced manner, thereby avoiding semantically unreasonable and inconsistent overlaps between new and old class regions that may arise from random placement. 
Moreover, EIR introduces a Region-Specific Knowledge Distillation loss (RSKD) to more effectively leverage the old-class instances that have been fused into the new images. This ensures consistent predictions between the old and new models in the old instance regions of the new images while avoiding adverse effects on the plasticity of the new classes. 
Our main contributions are three-fold:

    \begin{itemize}
    % \vspace{-5px}
    \item We show the limitations of existing image replay methods in CSS tasks, and this work is the first to explore a more suitable instance replay method.
    \item We discover that the two primary challenges in instance replay are dynamic category combination and background integration.
    \item We propose a novel instance replay framework (EIR) for CSS. EIR adeptly facilitates dynamic inter-class combinations of instances and integrates old-class instances with the background of new images, effectively addressing the issue of background shift in new images.
    \end{itemize}

\vspace{-0.1in}

\end{spacing}
\vspace{-0.025in}
\section{Related Work}
\vspace{-0.025in}
\begin{spacing}{0.96}

\noindent
\textbf{Continual learning}.
Continual learning focuses on enabling models to acquire new knowledge while preventing catastrophic forgetting~\cite{french1999catastrophic}, primarily studied in image classification. Existing methods fall into two groups: non-replay and replay~\cite{de2021continual}. Non-replay methods, such as architecture-based approaches~\cite{hu2023dense,yan2021dynamically,du2022agcn,du2023multi}, allocate specific parameters for each incremental task and progressively freeze and expand the model to mitigate forgetting.
Additionally, regularization-based methods~\cite{dhar2019learning,du2024confidence,du2024rebalancing} enhance retention of previous knowledge by incorporating regularization terms into the loss function, such as knowledge distillation~\cite{rebuffi2017icarl,simon2021learning} or gradient penalty~\cite{lyu2023measuring}. Replay-based methods~\cite{robins1995catastrophic,belouadah2019il2m,kim2021continual} utilize a fixed-size memory buffer~\cite{castro2018end,chaudhry2019tiny,hayes2020remind} or generative models~\cite{kemker2017fearnet,shin2017continual} during training to store, generate, and replay old images. In classification, replay effectively prevents forgetting since background shift is absent. However, in CSS, stored images suffer from partial labeling, limiting the storage advantage. 

% In classification tasks, the issue of background shift does not exist, making replay-based methods effective in alleviating catastrophic forgetting. However, in CSS, the stored images have the issue of partial labeling, which weakens the ability to fully utilize the storage advantage.

% \vspace{-0.1in}
\noindent
\textbf{Replay-free methods for CSS}.
% \vspace{-0.2in}
In recent years, CSS tasks have gained significant attention. MiB~\cite{cermelli2020modeling} was the first to highlight the issue of background shift, addressing it through background modeling. PLOP~\cite{douillard2021plop} preserved long-range and short-range spatial relationships at the feature level by applying feature distillation. SDR~\cite{michieli2021continual} employed prototype matching and contrastive learning to supplement existing knowledge distillation techniques. DKD~\cite{baek2022decomposed} utilized decomposed knowledge distillation to mitigate the forgetting of old classes, preserving both positive and negative output logits. SSUL~\cite{cha2021ssul} froze the model's backbone and mitigated the background shift issue by introducing unknown classes via a saliency detector. EWF~\cite{xiao2023endpoints} fused old and new knowledge with a weight fusion strategy.

%, and a significant number of methods have already been developed for this task
\begin{figure*}[t!]
	\centering
        \vspace{-0.15in}
\includegraphics[width=1.0\linewidth]{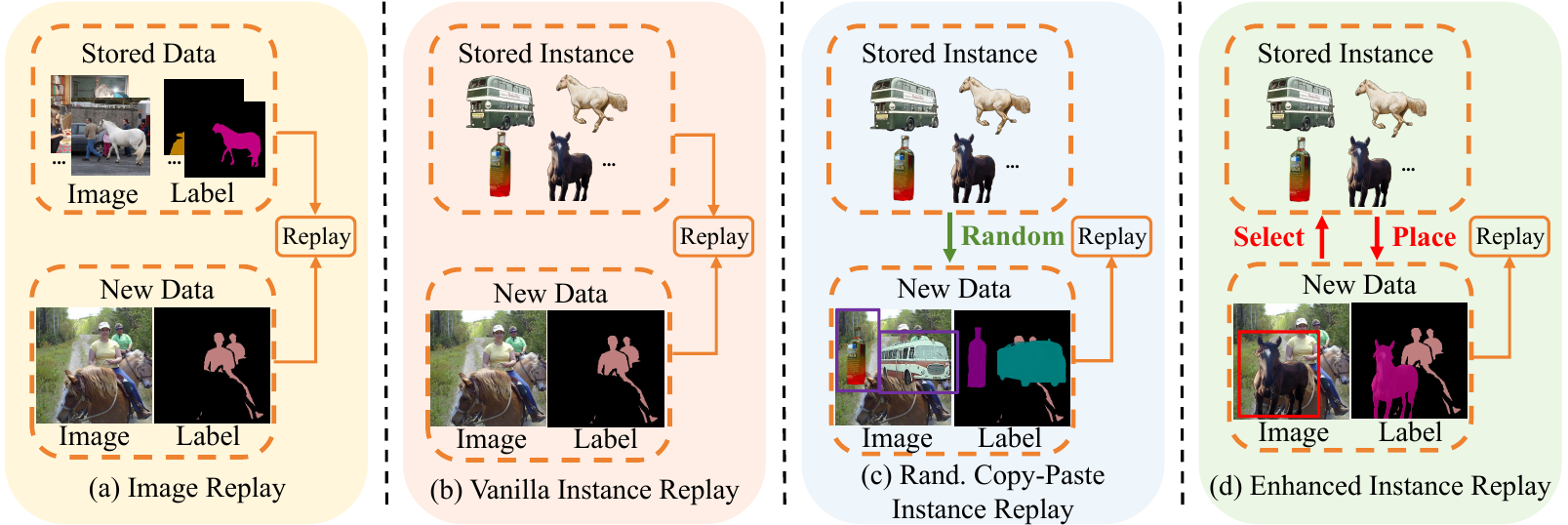}
	\vspace{-0.25in}
	\caption{Demonstration of four replay methods in CSS tasks. The figure above shows the detail implementation process of the image replay, vanilla instance replay, random copy-paste instance replay and enhanced instance replay. }
	\label{fig:22}
	\vspace{-0.1in}
\end{figure*}

% \vspace{-0.1in}
\noindent
\textbf{Replay-based methods for CSS}. RECALL~\cite{maracani2021recall} generated replay samples by crawling images from networks, albeit at a high computational cost or requiring additional data. \citet{zhu2023continual} improved performance using replay-based methods, filtering stored samples through a reinforcement learning paradigm. However, they neglected the confusing background information present in stored images, which hinders the full exploitation of replay's potential in preserving old knowledge. \yhm{PLOPLong \cite{douillard2021tackling} proposed a copy-paste strategy that randomly places old class instances into new images while removing the surrounding background. However, background removal will hinder the model’s understanding of class-environment relationships.} Summarizing the above thoughts, we design a replay method suitable for CSS tasks, which strategically places and augments old class instances in the background of new images.

\section{Rethinking Replay Strategy in CSS}
\subsection{Problem Definition of CSS}

Following \cite{cermelli2020modeling,douillard2021plop}, CSS aims to train the model through $T$ learning steps. At step \(t\), we present a dataset \(D_{t}\) consisting of pairs \(\left ( x_{t}, y_{t} \right  ) \), where \(x_{t}=\left \{ x_{t,i}  \right \}_{i=1}^{Q}\) denotes an input image of size \(H\times W\) with \(Q\) pixels, and \(y_{t}=\left \{ y_{t,i}  \right \}_{i=1}^{Q}  \) represents the corresponding ground truth segmentation mask. Specifically,  \(y_{t} \) consists only the labels from the new classes \(C_{t} \), while all other classes (i.e., old classes \(C_{1:t-1}\) and future classes \(C_{t+1:T} \)) are assigned to the background class \(c_{b}\). After the \(t\)-th step, the semantic segmentation model \(f_{\theta }^{t} \) is expected to predict whether a pixel belongs to any of the classes learned so far \(\left ( C_{1:t}=C_{1:t-1}\cup C_{t} \right  ) \) , or to the background class \(c_{b}\).

\subsection{Image replay for CSS}

For continual learning of image classification, image replay~\cite{lin2023pcr,chaudhry2019tiny} serves as a simple yet effective method. This method stores a small subset of data of previously learned classes and incorporates it with new data during the model's training, effectively mitigating the problem of catastrophic forgetting. However, in the CSS task, each stored image may simultaneously contain both new and old classes, {with annotations available only for partial old classes, while the remaining regions are labeled as background}.
Consequently, even when images containing old classes and their annotations are stored via image replay, \textit{the absence of corresponding annotations in stored and new images may result in confusion between foreground and background}, thereby limiting its effectiveness in mitigating catastrophic forgetting. This motivates the exploration of an enhanced method built upon image replay, aiming to effectively suppress background shift and alleviate catastrophic forgetting.

\vspace{-0.03in}
\subsection{Instance replay for CSS: two naive strategies}
\vspace{-0.03in}

As outlined above, the image replay strategy is ineffective for CSS due to the partial label challenge. 
Because of the challenge, stored images consist of fixed combinations of multiple instances, with only a subset labeled, which restricts the label space available during replay.
Since fixed combinations of partially-labeled instances can make the replay strategy ineffective, \textit{can we abandon the original combination method and instead store instances individually for dynamic combinations?} In the following, we describe two intuitive and direct approaches to instance replay.

\begin{figure}[t!]
	\centering
	\includegraphics[width=1.0\linewidth]{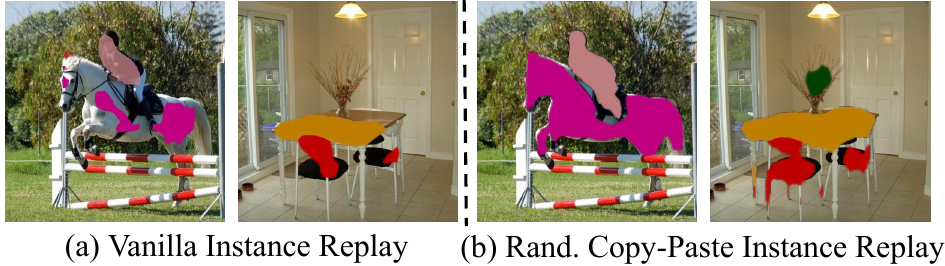}
	\vspace{-0.25in}
	\caption{Segmentation Results by Vanilla Instance Replay and the Random Copy-Paste Instance Replay. In the images above, the colored regions represent the model's predictions for the old classes, while the regions without additional color annotations indicate that the model predicted these areas as background. 
    \yhm{The corresponding numerical results are shown in Table~\ref{tab: replay comparison}.
    }
    }
	\label{fig:33}
	\vspace{-0.3in}
\end{figure}

\noindent
\textbf{Vanilla Instance Replay Mimicking Image Replay}~~
A simple way to implement instance replay is by following the image replay, which stores and retrains instances directly, as shown in Fig.~\ref{fig:22}. To evaluate the effectiveness, we conduct the instance replay experiment, with results shown in Fig.~\ref{fig:33}(a). The results show that there is a clear misclassification of old classes as background, suggesting that the model still significantly struggles to differentiate between old classes and the background. We believe that the performance of CSS is highly dependent on the inter-class relationships, which are often neglected when isolated, stored instances of old classes are directly input into the model for training. Moreover, the backgrounds of new images exhibit the issue of background shift, exacerbating the model's erroneous classification of old classes as background classes. 

\noindent
\textbf{Random Copy-Pasting Instance Replay}~~
Vanilla instance replay ignores the inter-class relationships between diverse old-class instances and is inadequate in addressing the issue of background shift in new images. Inspired by the advancement in recent data augmentation techniques~\cite{yun2019cutmix}, we then propose a simple random copy-pasting approach. This method involves randomly selecting old-class instances and integrating them into new images through a copy-paste process, followed by training the model on these augmented images. Compared to the vanilla method, this approach decreases the region where old classes are incorrectly classified as background as shown in Fig.~\ref{fig:33}(b). Nonetheless, the inherent randomness of the random copy-pasting process, particularly during the dynamic combination of classes and the integration of instances into new image backgrounds, will introduce noise and result in unstable training. For instance, as depicted in Fig.~\ref{fig:22}(c), the incorrect embedding of a ``bottle'' instance into an unrelated outdoor scene introduces a prominent issue of class inconsistency. 
Furthermore, in Fig.~\ref{fig:22}(c), the random copy-pasting strategy, due to its neglect of the paste location, exacerbates this issue. Consequently, the old class ``bus'' is inappropriately positioned within the new class ``person'', resulting in unreasonable and semantically inconsistent overlaps. This haphazard integration approach will impede the effective learning of both new and old classes, thereby undermining the anticipated efficacy of the instance replay. 

Vanilla instance replay is incapable of achieving dynamic class combination and effectively mitigating background shift in new images.  While random copy-paste instance replay can somewhat mitigate background shift, its randomness leads to noise, complicating the achievement of suitable dynamic class combinations. Moreover, there are risks involved in randomly integrating instances into the new image background, which can contaminate the training samples and limit the full potential of instance replay. 

% \vspace{-0.05in}

\end{spacing}
\vspace{-0.06in}
\section{Enhanced Instance Replay for CSS}
\begin{spacing}{0.95}

\subsection{Pipeline of the Enhanced Instance Replay}
\vspace{-0.02in}
In this paper, we propose an Enhanced Instance Replay (EIR) method to mitigate catastrophic forgetting and background shifts in the CSS task through replay. As discussed in the previous section, naive replay approaches overlook reasonable dynamic combinations across multiple classes and struggle to integrate effectively with the background. The proposed enhanced instance replay method addresses these two issues in a lightweight manner. From a practical perspective, we break down the implementation of EIR in the CSS task into several detailed steps: instance storage (Sec.~\ref{sec:storage}), class combination, instance selection (Sec.~\ref{sec:select}), instance placement (Sec.~\ref{sec:place}), and an enhanced replay training strategy (Sec.~\ref{sec:train}). Each of these components will be introduced in detail in the following sections. 

%4.1 整体的instance 怎么做？ 整体的流程是什么
%一上来就说存储，很突兀。 
%不一定完全按照  存取贴用，这样分 就太多了
%存：随机存   
%取：动态组合，贴什么  画成一个图 
%贴：背景融入，怎么贴 
%用：replay训练，在这里写我的 损失函数  在用这里写损失函数

%分几块，让大家看到 动态组合 和 背景融入到底是什么关系，这样到底有什么？
\vspace{-0.06in}
\subsection{Instance Storage for each CSS Task}
\vspace{-0.03in}
\label{sec:storage}

First of all, to replay at the instance level, we need to store instances for each CSS task.
% During instance storage, we necessitate the storage of instances across various classes like naive instance replay.
By definition, we denote \( \mathcal{M} \) as the memory buffer used to store instances of old classes. \( \mathcal{M} \) consists of instance pairs \(\left ( m_{p},m_{q}   \right )  \), where \( m_{p} \) represents the instance and \( m_{q} \) its label. At each step $t$, $\mathcal{M}$ maintains a balanced number of instances from all old classes $C_{1:t-1}$. Following the learning of class $C_{1:t}$ from training data $D_{t}$, we sample and store $\frac{|\mathcal{M}|}{|C_{1:t}|}$ instances per new class, giving preference to those containing a solitary, contiguous object region. To uphold equilibrium, we eliminate an equivalent number of instances from each old class upon the addition of a new one. Consequently, $\mathcal{M}$ consistently contains instances from all acquired classes $C_{1:t}$.

% \vspace{-0.2in}
\begin{figure*}[t!]
	\centering
        \vspace{-0.15in}
	\includegraphics[width=1.0\linewidth]{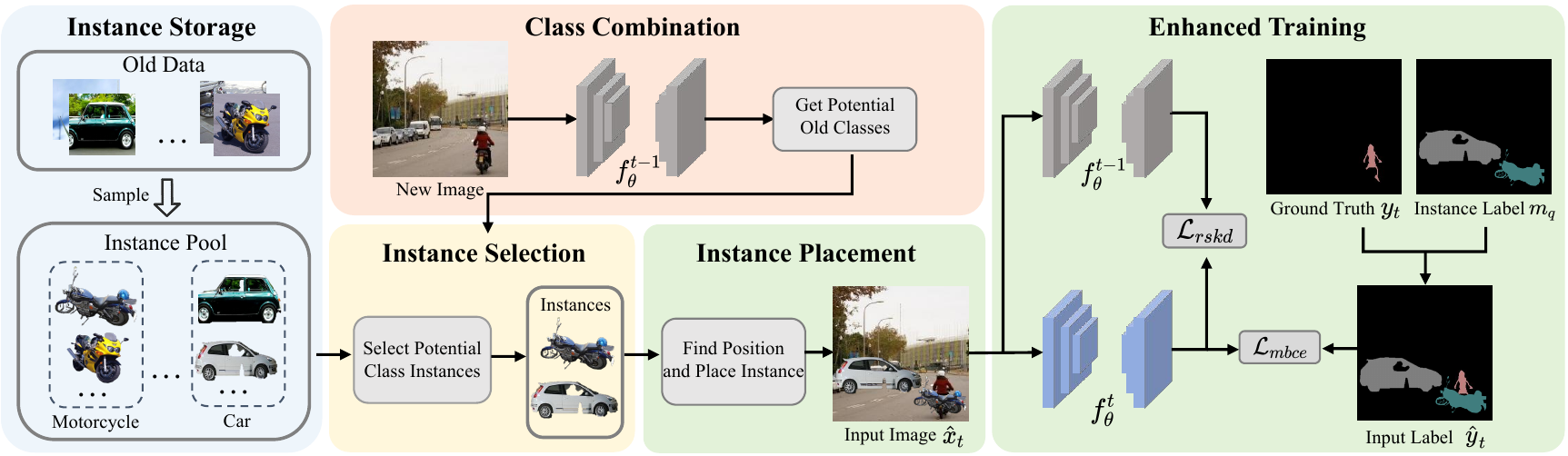}
	\vspace{-0.25in}
	\caption{The detailed architecture of our method. Initially, we sample instances from the old data according to their classes. Subsequently, during the class combination, we identify the potential old classes through the old model. In the instance selection, we select the instances of potential classes from the instance pool. After that, we calculate the position in new image to replay the instance and fuse them with the new image to create a fused image. Finally, the fused image is trained in an enhanced way.}
	\label{fig:3}
	\vspace{-0.2in}
\end{figure*}

\subsection{Class Combination and Instance Selection} 
\vspace{-0.005in}
\label{sec:select}
% 在训练新任务之前，我们已经获得了一组旧类别的实例。如前所述，即使来自同一新任务的图像在背景上也可能存在显著差异，错误地将不合适的实例与新任务图像结合进行回放训练可能会产生负面效果。因此，我们认为有必要选择合适的实例与每个训练图像进行结合，以确保在实例整合前后类别上下文的一致性。我们提出了一种基于原始图像上下文信息的实例组合策略。我们的策略包括以下几个实践。
Before training a new task, we have obtained a set of old class instances. As analyzed before, even images from a same new task can differ significantly in context, and improperly combining an unsuitable instance with a new task image for replay training could have negative effects. Therefore, we believe it is necessary to select appropriate instances to combine with each training image to ensure consistency in class context before and after instance integration. We propose an instance combination strategy based on the contextual information of the original image. Our strategy contains the following practices.

First, for each different image, because of the partial label issue, we need to find potential old classes within the image. \yhm{We directly utilize the old model to predict the class and corresponding logit value for each background pixel, considering those with logit values exceeding the threshold \(\tau\) as high-confidence old classes pixels \((\tau=0.7)\). Subsequently, we rank the potential old classes \( C \) based on the number of high-confidence pixels they contain.} These potential old classes may be contained in the image's context and serve as a basis for adjusting class combinations. We combine these potential old classes \( C \) with the new classes in each image. Then, we select instance samples from the instance buffer that corresponds to the chosen classes \( C \). 
In our implementation, we select the top two ranked class instances and each image is matched with a maximum of two instances, as the overlap condition imposes constraints on integration by limiting the number of instances that can fit without excessive overlap. Specifically, when multiple instances are introduced, they must occupy distinct spatial regions within the image to maintain clear boundaries between objects. If the overlap exceeds a certain threshold, it can lead to inaccurate instance blending and degrade the model’s learning effectiveness. Therefore, we restrict the number of instances to ensure each is distinguishable and effectively contributes to the model’s learning process.

\vspace{-0.07in}
\subsection{Instance Placement}
\label{sec:place}

Given the selected instances in the above subsection, we then need to place the instances into the new image.
\textit{The key question is where and how to place the instance.}
Because of the overlapping issue from in the instance placement, we develop a refined instance placement strategy. 

To locate where to place and avoid potentially confusing information in the new image background, correctly placing old class instances is important. 
If an old class instance overlaps excessively with a new class region, it not only disrupts the effective learning of new classes but may even hinder it. 
To address this issue, we propose placing old class instances within the image background as much as possible to minimize their intersection with new class regions. 
Specifically, we first divide the image into $n$ \yhm{rectangular} regions. 
For each region, we calculate the proportion of background pixel it contains and select the top-left corner of the region with the highest proportion as the coordinate \((u, v)\). If there are multiple regions with the same proportion, we choose the one closest to the image's top-left corner. This area then determines the initial top-left coordinate for placing the old class instance. Then, we align the top-left corner of the old class instance with \((u, v)\) and resize it as needed to ensure the entire instance fits within the image boundaries. For each pixel in the old class instance, we place it to the corresponding pixel in the new image at position \((\hat{u}, \hat{v})\), where $\hat{u} = u + a$ and $\hat{v} = v + b$, with $a$ and $b$ representing the relative positional offsets within the old class instance. 

%原话：If there are multiple regions with the same proportion, we choose the one closest to the top-left corner of the image. 

After aligning the top-left corner of the instance with $(u, v)$, we apply the mixup technique to fuse the old class instance \(\left ( x_{t},y_{t} \right ) \) into the new image in a coherent manner:
\begin{equation}
\left\{
\begin{aligned}
\hat{x}_{\mathrm{t}}(\hat{u}, \hat{v}) &= \lambda x_{\mathrm{t}}(\hat{u}, \hat{v}) + (1-\lambda) m_{p}(a, b), \\
\hat{y}_{\mathrm{t}}(\hat{u}, \hat{v}) &= \lambda y_{\mathrm{t}}(\hat{u}, \hat{v}) + (1-\lambda) m_{q}(a, b).
\end{aligned}
\right.
\end{equation}
\( \lambda \) is a value sampled from beta distribution within the range $[0,1]$, controlling the blending between the old class instances and new image regions. 
\yhm{The mixup technique enhances class-background distinction through linear interpolation, creating smoother transitions between fused images.} 

%\vspace{-3mm}
\subsection{Enhanced Replay Training Strategy}

\label{sec:train}

Traditional image replay methods typically employ knowledge distillation loss during training, which enforces prediction consistency between the old and new models when the pixel belongs to the new class region.
This may hinder the model's ability to learn new classes, as the old model has not learned the new class, thus failing to fully leverage the benefits of replaying old class instances within new images. 
Thus, we design a Region-Specific Knowledge Distillation loss (RSKD), to selectively target and train these distinct regions within the images, which is formulated as:
\vspace{-0.05in}
\begin{equation}
\!\!\!\scalebox{0.95}{$
\mathcal{L}_{rskd} = \frac{1}{HW} \left\{
\begin{aligned}
&\sum_{i=1}^{HW} \sum_{c \in A} p_{i,c}^{t-1} \log p_{i,c}^{t}, ~~~~~~~~~~~~\text{if } c_{i} \in A, \\
&\sum_{i=1}^{HW} p_{i,c_{b} }^{t-1} \log \left( \sum_{k \in B \cup c_{b}} p_{i,k}^{t}  \right),  \text{if } c_{i} \in B,
\end{aligned}
\right.
$}
% \tag{4}
\end{equation}
where \(A \) denotes the set of old classes and the background class \(\ C_{1:t-1} \cup  c_b \), and \(B \) denotes the set of the new classes \(\ C_t \). \(p_{i,c_{b} }^{t-1} \) is the probability predicted by old model \(f_{\theta }^{t-1} \) as background and \(p_{i,k}^{t} \)  is the probability predicted by new model \(f_{\theta }^{t} \) as class \(k\). In the RSKD loss, if the pixel belongs to old classes and background region, we align the prediction of the old model \(f_{\theta }^{t-1} \) and new model \(f_{\theta }^{t} \), leveraging the old model’s acquired knowledge of old classes. If the pixel \(i\) belongs to the new class region, RSKD loss encourages \(p_{i,c_{b} }^{t-1} \) similar to the sum of \(p_{i,k}^{t} \) for \(k \in  C_{t} \cup c_{b}\). 

Moreover, our class combination approach relies on predicting potential old classes within images and that some old classes may have a low probability of appearing in the image background, we implement a training strategy where, while integrating old class instances into new images, we input each old class instance into the model for training just like the vanilla instance replay. \( m_{q} \) represents the ground truth corresponding to the instance, and \(y_{q}\) represents the model's prediction for the instance. \( \hat{y} \) represents the ground truth corresponding to the image, and \(\hat{y}_{p}\) represents the model's prediction for the image.
% \begin{equation}
%     \mathcal{L}_{\text {vanilla }}=\mathcal{L}_{\text {mbce }}\left ( m_{q} , y_{q} \right ) 
% \end{equation}
This dual approach not only enhances the dynamism of class combination but also ensures effective integration and improves the overall robustness and adaptability of our method.
We formulate the overall training objective as:
\vspace{-0.05in}
\begin{equation}
    \mathcal{L}_{\text {total}}=\mathcal{L}_{\text {mbce }}\left ( m_{q} , y_{q} \right ) + \mathcal{L}_{\text {mbce }}\left ( \hat{y} , \hat{y}_{p} \right ) + \alpha \mathcal{L}_{rskd}.
\end{equation}
% \vspace{-0.2in}
% \begin{align}
% \scalebox{1}{$
% \mathcal{L}_{total} = \mathcal{L}_{vanillla} + \mathcal{L}_{image}.
% $}
% \end{align}
% \begin{align}
% \scalebox{1}{$
% \mathcal{L}_{total} = \mathcal{L}_{vanilla} + \mathcal{L}_{mbce} + \alpha \mathcal{L}_{rskd}.
% $}
% % \tag{5}
% \end{align}
Here, $\mathcal{L}_{mbce}$ denotes multiple binary cross-entropy (mBCE) loss, \(\alpha\) is the hyper-parameter to balance two loss terms.

% If the class of pixel \(i\) belongs to old classes \(C_{1:t-1} \) or background \(c_{b}\), we align the prediction of the old model \(f_{\theta }^{t-1} \) and new model \(f_{\theta }^{t} \) to mitigate forgetting, leveraging the old model’s acquired knowledge of old classes. However, enforcing prediction consistency between the old and new models when the class of pixel \(i\) belongs to new classes \(C_{t} \) may hinder the model's ability to learn new classes, as the old model has not learned the new class \(C_{t} \). Therefore, if the class of pixel \(i\) belongs to new classes \(C_{t} \), RSKD loss encourages \(p_{i,c_{b} }^{t-1} \) similar to the sum of \(p_{i,k}^{t} \) for \(k \in  C_{t} \cup c_{b}\), where \(p_{i,c_{b} }^{t-1} \) is the probability predicted by old model \(f_{\theta }^{t-1} \) as background and \(p_{i,k}^{t} \)  is the probability predicted by new model \(f_{\theta }^{t} \) as class \(k\). 

\end{spacing}
\vspace{-0.1in}
\section{Experiments}
\begin{spacing}{0.95}

\subsection{Expermental Setup}

%\subsubsection{Datasets}
{\bf Datasets~~} We evaluate our method on datasets of Pascal VOC 2012\cite{everingham2010pascal} and ADE20K\cite{zhou2017scene}, which are the standard CSS benchmarks. Pascal VOC 2012 comprises 10,582 training images and 1,449 validation images, covering 20 classes. ADE20K includes 20,210 training images and 2,000 validation images, spanning 150 classes.

\noindent
%\subsubsection{Experimental Setting}
{\bf Experimental Setting~~} We established two experimental protocols: disjoint and overlapped. The distinction between these setups lies in the presence of future classes \(C_{t+1:T}\). In the disjoint setting, the images do not include future classes, whereas in the overlapped setup, they do, making the latter more challenging and reflective of real-world scenarios. Consequently, our experiments focus on overlapped CSS, with additional results for disjoint CSS provided in the supplementary materials. We evaluated several CSS protocols for each dataset. For the Pascal VOC 2012, we considered common experimental settings such as 15-5 (2 steps), 5-3 (6 steps), and 15-1 (6 steps), as well as more challenging setups like 10-1 (11 steps) and 2-2 (10 steps), which involve more incremental steps and closely mimic real-world scenarios. For the ADE20K, we conducted experiments under three settings: 100-50 (2 steps), 100-10 (6 steps), and similar incremental configurations.

\noindent
%\subsubsection{Metrics}
{\bf Metrics} We utilize the mean Intersection-over-Union (mIoU) for evaluation. To comprehensively assess the model's various performances in the context of CSS, we report the average IoU scores for the initial classes, incremental classes, and all classes. These metrics respectively reflect the model's robustness against catastrophic forgetting (rigidity), its ability to learn new classes (plasticity), and its overall performance, which balances rigidity and plasticity.

\noindent
%\subsubsection{Implementation Details}
{\bf Implementation Details~~} Following previous work, we utilize DeepLabv3~\cite{chen2017rethinking} with a ResNet-101~\cite{he2016deep} backbone pre-trained on ImageNet~\cite{deng2009imagenet} as our segmentation network. We also conduct experiments using the Transformer framework and choose Swin Transformerbase (Swin-B) ~\cite{liu2021swin} as the backbone. As in \cite{douillard2021plop,cha2021ssul}, we employ different training strategies for the two datasets. For Pascal VOC 2012, we train for 60 epochs with an initial learning rate of 0.001 for the first step and 0.0001 for incremental steps, empirically setting \(\gamma\) to 4. For ADE20K, we train for 100 epochs with an initial learning rate of 0.00025 for the first step and 0.000025 for incremental steps, setting \(\gamma\) to 30. For both datasets, we use an SGD optimizer with a momentum of 0.9. The batch size is set to 24, with \(\alpha\) and \(\beta\) set to 5 and 0.05, respectively. Our model is implemented using PyTorch toolkit on one NVIDIA RTX 4090 GPU for acceleration.

\begin{figure*}[t!]
    \centering
    \includegraphics[width=1\linewidth]{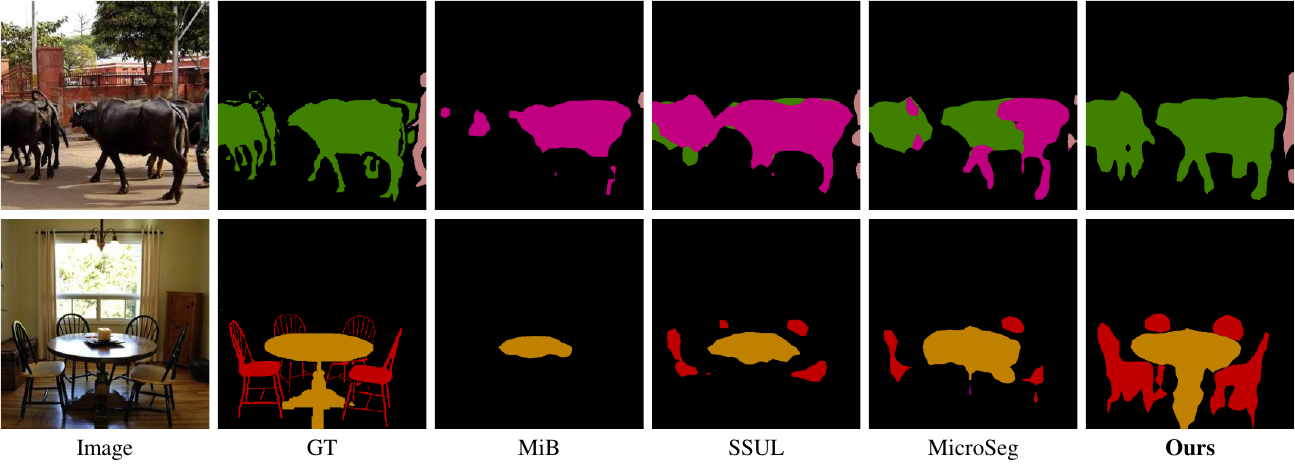}
    \vspace{-0.35in}
    \caption{Segmentation results of our method and previous methods on Psacal VOC 2012.}
    \label{fig: visual}
    \vspace{-0.05in}
\end{figure*}

% \vspace{-0.1in}

\subsection{Ablation Study}
% \vspace{-0.05in}

\begin{table}[t]
    \centering
    \caption{Results of different replay strategies. 
    The performance is reported on Pascal VOC 5-3 overlapped setting.}
   % ？ \vspace{0.05in}
    \vspace{-0.05in}
    \renewcommand\tabcolsep{8.7pt}
    \renewcommand{\arraystretch}{1.1}
    % \vspace{0.05in}
    \begin{tabular}{c|ccc}
    \Xhline{1.3pt}
    \multirow{1}*{Method} & 0-5 & 6-20 & all\\
    \hline      
    % none &74.59 & 53.16 & 58.28\\ 
    Image replay & 55.36 & 37.34 & 42.49\\ 
    Vanilla instance replay & 52.70 & 53.31 & 53.14\\
    Random copy-pasting & 59.72 & 56.35 & 57.30\\
    Enhanced instance replay & {\bf 74.61} & {\bf 63.13} & {\bf 66.41}\\ 
    \Xhline{1.3pt}
    \end{tabular}
    \label{tab: replay comparison}
    \vspace{-0.1in}
\end{table}

\begin{table}[t]
    \centering
    \caption{Analysis of Our Method’s Components, tested on 5-3 overlapped setting of Pascal VOC.}
    \renewcommand\tabcolsep{3.5pt}
    \renewcommand{\arraystretch}{1.1}
    \vspace{-0.1in}
    \begin{tabular}{ccc|ccc}
    \Xhline{1.3pt}  
    \multirow{1}*{Combination} & \multirow{1}*{Placement} & \multirow{1}*{Enhance} & 0-5 & 6-20 & all\\
    \hline
    \checkmark & &  & 66.02 & 57.58 & 59.99\\ 
    &\checkmark& & 61.99 & 57.14 & 58.52\\ 
    \checkmark& \checkmark & & 70.16 & 60.36	& 63.16\\ 
    & & \checkmark & 65.53 & 56.78 & 59.28  \\ 
    \checkmark & \checkmark &\checkmark& {\bf 74.61} & {\bf 63.13} & {\bf 66.41}\\
    \Xhline{1.3pt}
    \end{tabular}
    \vspace{-0.1in}
    \label{tab: component}
    \vspace{-0.1in}
\end{table}

\noindent
{\bf Comparison of Replay Strategies~~} In Table~\ref{tab: replay comparison}, we present a Comparison on four different replay strategies, including one image replay method and three distinct forms of instance replay. The image replay (see ``Image replay'') mitigates catastrophic forgetting by storing and utilizing images containing old classes. However, the presence of background shift in both old and new class data results in the poorest performance. 
While vanilla instance replay (see ``Vanilla instance replay'') addresses the background shift issue present in image replay by storing instance of old classes, it fails to resolve the background shift in images of new class. 
The random copy-pasting (see ``Random copy-pasting'') improves vanilla instance replay by pasting stored instance of old classes on images of new class. However, the randomness of the pasting process limits its effectiveness in mitigating the background shift. In contrast, Our approach (see ``Enhanced instance replay'') effectively mitigates background shift issues across both new and old class data through instance combination and placement, ultimately yielding superior results.

\noindent
{\bf Analysis of Our Method’s Components~~}In Table~\ref{tab: component}, we conduct an analysis of the individual components in our method. Initially, based on random copy-pasting, we performed dynamic combination of old classes (see first row). Although this approach allows us to selectively paste instances from certain categories onto new images, the lack of positional context information limits the model's segmentation capability. Next, we employed ordered pasting of randomly selected instances onto new images (see second row). While this method partially addresses the positional context of different categories, the overwhelming presence of out-of-context instances in image disrupts the model's ability to learn semantic context. Additionally, we separately introduced an enhancement strategy within replay training framework (see fourth row). Without a more refined instance replay strategy, the enhancement only yielded limited improvements. Finally, when all three strategies are combined (see last row), the best performance is achieved.

\begin{table}[t]
    \centering
    \vspace{-0.15in}
    \caption{Results of different instance combination strategies on Pascal VOC 5-3 overlapped setting.}
    \renewcommand\tabcolsep{14.8pt}
    \renewcommand{\arraystretch}{1.1}

    \vspace{-0.1in}
    \begin{tabular}{c|ccc}
    \Xhline{1.3pt}
    \multirow{1}*{Method} & 0-5 & 6-20 & all\\
    \hline      
    % none &74.59 & 53.16 & 58.28\\ 
    Random & 74.08 & 60.78 & 64.58\\ 
    Class-order & 73.22 & 60.37 & 64.04\\
    Pseudo-label & {\bf 74.61} & {\bf 63.13} & {\bf 66.41}\\ 
    \Xhline{1.3pt}
    \end{tabular}
    \vspace{-0.05in}
    \label{tab: combination}
    % \vspace{-0.15in}
\end{table}

\begin{table}[t]
    \centering
    % \vspace{-0.0in}
    \caption{Results of different instance placement strategies. The performance is reported on Pascal VOC 5-3 overlapped setting.}
    \renewcommand\tabcolsep{11.5pt}
    \renewcommand{\arraystretch}{1.1}
    \vspace{-0.1in}
    \begin{tabular}{c|ccc}
    \Xhline{1.3pt}
    \multirow{1}*{Method} & 0-5 & 6-20 & all\\
    \hline
    No-placement & 47.05 & 43.83 & 44.75\\ 
    Random paste & 72.60 & 59.69 & 63.38\\  %
    Strategic-placement & {\bf 74.61} & {\bf 63.13} & {\bf 66.41}\\ 
    \Xhline{1.3pt}
    \end{tabular}
    \vspace{-0.2in}
    \label{tab: placement}   
    % \vspace{-0.05in}
\end{table}

\noindent
{\bf Analysis of Instance Combination~~} In Table~\ref{tab: combination}, we analyze the different strategies for selecting class of instance for each new image. We perform the following experiments to validate the effectiveness of our method: (1) Random: randomly selecting instance for combination (see ``Random''), (2) Class-order: determining instance classes based on class order (see ``Class-order''), and (3) Pseudo-label: selecting instance classes based on newly added classes identified in pseudo labels. In the aforementioned approach, the pseudo-label strategy yielded superior results, particularly in enhancing the performance on new classes. We attribute this success to the method’s ability to maintain contextual consistency within the images, enabling the model to accurately learn the context of both new and existing classes. This not only strengthens the learning of new classes but also mitigates the forgetting of previously learned ones.

\noindent
{\bf Analysis of Instance Placement~~} In Table~\ref{tab: placement}, we also investigated the impact of different strategies of instance placement. Here, we compared three strategies: (1) No-placement: involving no instance placement, (2) Random paste: which randomly pastes instances into the background of new data, and (3) Strategic-placement, which employs our instance placement strategy to fuse instances into the new data’s background. The comparison between ``Random paste'' and ``Strategic-placement'' reflects that the our placement strategy can enhance the spatial context for model training. Additionally, we observed that while the overall performance of the random paste is better than No-placement, the performance for old classes is inferior. This supports our previous assertion that random instance replay can potentially lead to negative impacts and further underscores the necessity of using the Strategic-placement strategy to ensure model stability and effectiveness.

\begin{table}[t]
    \centering
    \vspace{-0.15in}
    \caption{Performance on the Pascal VOC 5-3 overlapped setting when using different enhanced replay training strategies.}
    \renewcommand\tabcolsep{14pt}
    \renewcommand{\arraystretch}{1.1}

    \vspace{-0.1in}
    \begin{tabular}{c|ccc}
    \Xhline{1.3pt}
      \multirow{1}*{Method} & 0-5 & 6-20 & all\\
      \hline
      UNKD~\cite{cermelli2020modeling} & 66.90 & 49.88 & 54.74 \\ 
      DKD~\cite{baek2022decomposed} & 70.08 & 59.00 &62.16 \\ 
      RSKD~(Ours) & {\bf 74.61} & {\bf 63.13} & {\bf 66.41}\\ 
    \Xhline{1.3pt}
    \end{tabular}
    \vspace{-0.2in}
    \label{tab: kd_loss}
\end{table}

\begin{table*}[t]
    \centering
    \vspace{-0.15in}
    \caption{The mIoU of the last step on the Pascal VOC 2012 dataset for different continual semantic segmentation settings}
    \renewcommand\tabcolsep{3pt}
	\renewcommand{\arraystretch}{1.1}
    
    \vspace{-0.1in}
    \begin{tabular}{c|c|ccc|ccc|ccc|ccc|ccc}
    \Xhline{1.3pt}
    \multirow{2}*{\textbf{Method}} & \multirow{2}*{\textbf{Backbone}} & \multicolumn{3}{c|}{\textbf{VOC 10-1} (11)}&\multicolumn{3}{c|}{\textbf{VOC 15-1} (6)}& \multicolumn{3}{c|}{\textbf{VOC 5-3} (6)}& \multicolumn{3}{c|}{\textbf{VOC 2-2} (10)} & \multicolumn{3}{c}{\textbf{VOC 15-5} (2)}\\
    
    \multicolumn{1}{c|}{} &\multicolumn{1}{c|}{} & 0-10 & 11-20 & all & 0-15 & 16-20 & all & 0-5 & 6-20 & all & 0-2& 3-20& all & 0-15 & 16-20 & all \\
    \hline
    \hline
    Joint & Resnet101 &78.4 & 76.4&77.4&79.8 & 72.4 &77.4&76.9 & 77.6 &77.4&76.5 & 77.5 &77.4&77.8 & 72.4 &77.4\\
    ILT~\cite{michieli2019incremental}& Resnet101 & 7.2&  3.7&  5.5&  8.8&  8.0&  8.6&  22.5&  31.7&  29.0&  5.8&  5.0&  5.1&  67.1&  39.2&60.5\\
    MiB~\cite{cermelli2020modeling}& Resnet101& 12.3 & 13.1 & 12.7& 34.2 & 13.5 & 29.3 & 57.1 & 42.6 & 46.7 & 41.1& 23.4& 25.9& 76.4 & 50.0 &70.1\\
    PLOP~\cite{douillard2021plop}& Resnet101 & 44.0 & 15.5 & 30.5 & 65.1& 21.1& 54.6& 17.5& 19.2& 18.7 & 24.1& 11.9& 13.7& 75.7 & 51.7 &70.1\\
    SSUL-M~\cite{cha2021ssul} & Resnet101 & 72.4 & 49.8 & 61.6& 76.6 & 37.4& 67.3& 71.5 & 51.5& 57.2 & 62.4& 42.5& 45.3& 77.9& 55.6 &72.6\\
    % SSUL~\cite{cha2021ssul} & Resnet101 & 71.3 & 46.0 & 59.3& 77.3 & 36.6& 67.6& 72.4 & 50.7& 56.9& 62.4& 42.5& 45.3& 77.8& 50.1 &71.2\\
    MicroSeg-M~\cite{zhang2022mining} & Resnet101 & \textbf{73.1}& 52.9& 63.5& \textbf{80.4}& 42.2& 71.3& 73.5& 60.3& 64.1& 59.4& 46.7& 48.5& \textbf{81.5}& 55.0&\textbf{75.2}\\
    RCIL~\cite{zhang2022representation} & Resnet101 & 55.4& 15.0& 36.2& 70.6& 27.4& 59.4& 55.4& 15.1& 34.3& -& -& -& 78.8& 52.0&72.4\\
    EWF~\cite{xiao2023endpoints}& Resnet101 & 71.5& 30.3& 51.9& 77.7& 32.7& 67.0& 69.0& 45.0& 51.8& -& -& -& -& -&-\\
    TIKP~\cite{yu2024tikp}& Resnet101 & 69.7&43.5 &57.2 & 73.8&42.3 &66.3 & -& -& -&-&-&-&78.8 &55.5 &73.3\\ 
    \hline
    {\bf Ours}& Resnet101 & 71.9 &{\bf 58.1} & {\bf 65.3} & 79.4& {\bf 52.6}& {\bf 73.0}&{\bf 74.6} &{\bf 63.1} &{\bf 66.4} &{\bf 68.3}&{\bf 55.0}&{\bf 56.9}&79.1 &{\bf 58.4} & 74.2\\
    \hline
    \hline
    Joint & Swin-B & 82.4 & 83.0 & 82.7 & 83.8 & 79.3 & 82.7 & 82.6 & 84.4 & 82.7 & 75.8 & 83.9 & 82.7 & 83.8 & 79.3 & 82.7  \\
    SSUL~\cite{cha2021ssul} & Swin-B & 74.3 & 51.0 & 63.2 & 78.1 & 33.4 & 67.5 &-&-&-&  60.3 & 40.6 & 44.0 & 79.7 & 55.3 & 73.9 \\
    MicroSeg~\cite{zhang2022mining} & Swin-B & 73.5 & 53.0 & 63.8 & 80.5 & 40.8 & 71.0 &-&-&-& 64.8 & 43.4 & 46.5 & 81.9& 54.0& 75.2\\
    CoinSeg~\cite{zhang2023coinseg} & Swin-B & \textbf{80.1} & 60.0 & 70.5 & 82.7 & 52.5 & 75.5 &-&-&- & 70.1 & 63.3 & 64.3 & 82.1 & 63.2 & 77.6  \\
    \hline  %之前的10-1的结果：{\bf 69.5} &{\bf 71.8} & {\bf 70.6} 
    {\bf Ours}& Swin-B & 75.9 &{\bf 71.3} & {\bf 73.7} & {\bf 83.6}& {\bf 66.9}& {\bf 79.6}&{\bf 74.5} &{\bf 73.00} &{\bf 73.4} &{\bf 70.9}&{\bf 69.2}&{\bf 69.5}&{\bf 83.4} &{\bf 68.6} &{\bf 79.9}\\
    \Xhline{1.3pt}
    \end{tabular}
    % \label{table1}
    \label{tab: VOC}
    \vspace{-0.05in}
\end{table*}

%\subsubsection{Baselines}
\begin{table*}[t]
    \centering
    \caption{The mIoU of the last step on the ADE20K dataset for different continual class segmentation settings.}
    \renewcommand\tabcolsep{8.2pt}
    \renewcommand{\arraystretch}{1.05}
    
    \vspace{-0.1in}
    \begin{tabular}{c|c|ccc|ccc|ccc}
    \Xhline{1.3pt}
    \multirow{2}*{\textbf{Method}} & \multirow{2}*{\textbf{Backbone}} & \multicolumn{3}{c|}{\textbf{ADE 100-50} (2)}&\multicolumn{3}{c|}{\textbf{ADE 50-50} (3)}& \multicolumn{3}{c}{\textbf{ADE 100-10} (6)}\\
    
    \multicolumn{1}{c|}{}& \multicolumn{1}{c|}{}& 0-100 & 101-150 & all & 0-50 & 51-150 & all & 0-100 & 101-150 & all \\
    \hline
    \hline
    Joint& Resnet101 &	44.3 &	28.2 &	38.9 &	51.0 &	33.3 &  38.9 &	44.3 &	28.2 &38.9\\ 
    ILT~\cite{michieli2019incremental}& Resnet101 &  18.3&  14.4&  17.0&  3.5&  12.9&  9.7&  0.1&  3.1& 1.1\\
    MiB~\cite{cermelli2020modeling}& Resnet101 & 40.5& 17.2& 32.8& 45.6& 21.0& 29.3& 38.2& 11.1&29.2\\
    PLOP~\cite{douillard2021plop}& Resnet101 & 41.8& 14.5& 32.7& 47.3& 20.3& 29.4& 38.6& 14.2&30.5\\
    SSUL-M~\cite{cha2021ssul}& Resnet101 &	41.3&	18.0&	33.6&	48.4&	20.2&   29.7&  41.3&  17.5&33.4\\ 
    MicroSeg-M~\cite{zhang2022mining}& Resnet101 &	40.4&	18.6&	33.2 &	48.3&	24.6&   32.6&  41.2&  {\bf 20.9}&{\bf 34.5}\\
    RCIL~\cite{zhang2022representation}& Resnet101 & {\bf 42.3}& 18.8& 34.5& 48.3& 25.0& 32.5& 39.3& 17.6&32.1\\
    EWF~\cite{xiao2023endpoints}& Resnet101 & 41.2& 21.3& 34.6& -& -& -& 41.5& 16.3&33.2\\ 
    TIKP~\cite{yu2024tikp}& Resnet101 &	42.2&	20.2&	34.9&	48.8&	25.9&   33.6&  41.0&  19.6&33.8\\ 
    \hline
    {\bf Ours}& Resnet101 &	41.9&	{\bf 21.9}&	{\bf 35.3}&	{\bf 49.3}&	{\bf 26.3}& {\bf 34.1}&	{\bf 41.8}&	19.4&34.3\\ 
    \hline
    \hline
    Joint & Swin-B & 43.5 & 30.6 & 39.2 & 50.2 & 33.7 & 39.2 & 43.5 & 30.6 & 39.2 \\
    SSUL~\cite{cha2021ssul} & Swin-B & 41.9 & 20.1 & 34.6 & 49.5 & 21.3 & 30.7 & 40.7 & 19.0& 33.5\\
    MicroSeg~\cite{zhang2022mining} & Swin-B & 41.1 & 24.1 & 35.4 & 49.8 & 23.9 & 32.5 & 41.0 & 22.6 & 34.8\\
    CoinSeg~\cite{zhang2023coinseg} & Swin-B & 41.6 & 26.7 & 36.6 & 49.0 & \textbf{28.9} & 35.6 & 42.1 & \textbf{24.5} &\textbf{ 36.2} \\
    \hline
    {\bf Ours}& Swin-B & {\bf 42.1} &{\bf 27.3} & {\bf37.2} & {\bf 49.7}& 28.8& {\bf 35.8}&{\bf 42.3} & 23.6 &36.1\\
    \Xhline{1.3pt}
    \end{tabular}

    \label{tab: ADE}
    \vspace{-0.15in}
\end{table*}

\noindent
{\bf Analysis of Enhanced Replay Training Strategy~~} In Table~\ref{tab: kd_loss}, we evaluate the constraint performance of different forms of knowledge distillation loss on the results obtained by the new and old models. Specifically, we replaced the proposed RSKD loss in our method with the unbiased knowledge distillation (UNKD)~\cite{cermelli2020modeling} and the decomposed knowledge distillation (DKD)~\cite{baek2022decomposed}. The results indicate that RSKD outperforms the other two KD losses in terms of constraint effectiveness. This superiority is attributed to RSKD's consideration of instances adhered to new class images, where it applies distillation constraints specifically to the regions involved in the mixup operation within the new class images.

\subsection{External Comparison}
In CSS tasks, stored images during replay are annotated with only a subset of old classes. However, SSUL-M~\cite{cha2021ssul} and Microseg-M~~\cite{zhang2022mining} deviate from the protocol by integrating all old class annotations when replaying old data. To accurately assess these methods, we refined experiments with SSUL-M and MicroSeg-M, aligning setups with CSS task specifications. In Tables~\ref{tab: VOC} and~\ref{tab: ADE}, we compare the performance of the most advanced CSS methods. It is clear that our approach outperforms all other methods in widely settings, such as the 15-1, 5-3, and 15-5 setups on Pascal VOC 2012 and the 100-50 and 50-50 setups on ADE20K, and also demonstrates a significant advantage in more challenging setups, such as the 2-2 and 10-1 setups on Pascal VOC 2012 and the 100-10 setup on ADE20K. \yhm{Note that although Microseg-M slightly outperforms on old classes in  scenarios with more base classes (VOC 15-5, VOC 15-1, VOC 10-1), it fails to effectively learn new classes,  leading to inferior overall performance compared to our approach. Moreover, in more challenging scenarios (VOC 2-2, VOC 5-3), its performance drops sharply, highlighting the better learning balance of our method on both new and old classes.} In Figures~\ref{fig: visual}, we compare the segmentation results of different methods, where our method yields better results. 

%问题：with more base classes   还是 in more-base-classes scenarios 
%旧版本：On the VOC dataset with ResNet backbone, Microseg-M slightly outperforms on old classes in settings with more old classes (15-5, 15-1, 10-1) but lags behind us in new classes and overall performance. In more challenging scenarios (2-2, 5-3), its performance drops sharply, highlighting our method's better balance between plasticity and stability.

%新版本：Note although Microseg-M outperforms on old classes in more-base-classes scenarios (VOC 15-5, VOC 15-1, VOC 10-1) when the backbone is ResNet, it fail to sufficiently address new class learning, resulting in inferior overall performance compared to out approach. Moreover, in more challenging scenarios (2-2, 5-3), its performance drops sharply, highlighting our method's better balance the learning of both new and old classes. Specifically, strategic instance placement prevents overlap between the new and old class regions, minimizing interference with new class learning.

\vspace{-0.05in}
\section{Conclusion}
\vspace{-0.05in}
This paper proposed an  enhanced instance replay method for CSS tasks. Based on instance storage, we integrated instances with new class images, efficiently alleviated the background shift issue. Therefore, our method can ultimately achieve superior segmentation results across both new and previously learned classes. Although our method effectively mitigate background shift issue, it does not fully resolve this problem because it cannot accurately cover all background shift regions in new images using instances, necessitating further exploration in future research.

%This work was supported in part by the National Key R&D Program of China under Grants 2023YFF0906200.
% \vspace{0.35in}
\noindent
{\bf Acknowledgment~~} This work was supported in the National Key R\&D Program of China under Grant No. 2023YFF0906200, the National Natural Science Foundation of China (Nos. 62276182, 62476196, 62406323), Peng Cheng Lab Program (No. PCL2023A08), Tianjin Natural Science Foundation (Nos. 24JCYBJC01230, 24JCYBJC01460), and Tianjin Municipal Education Commission Research Plan (No. 2024ZX008), China Postdoctoral Science Foundation (No. 2024M753496), and Postdoctoral Fellowship Program of CPSF (No. GZC20232993).

\end{spacing}

{
    \small
    \bibliographystyle{ieeenat_fullname}
    \bibliography{main}
}


% {
%     \small
%     \bibliographystyle{ieeenat_fullname}
%     \bibliography{main}
% }
\end{document}